\documentclass[letterpaper]{article}
\usepackage{aaai2027}
\nocopyright     %
\usepackage[hyphens]{url}
\usepackage{graphicx}
\usepackage{natbib}
\usepackage{caption}
\usepackage{algorithm}
\usepackage{algorithmic}
\usepackage{multirow}
\usepackage{booktabs}
\usepackage{tabularx}
\usepackage{amsmath}
\usepackage{stmaryrd}

\usepackage{newfloat}
\usepackage{listings}
\usepackage{xspace}
\usepackage{tikz}
\usetikzlibrary{arrows.meta,positioning,patterns,shapes.geometric}
\usepackage{pgfplots}
\pgfplotsset{compat=1.18}
\usepgfplotslibrary{groupplots}
\usepackage{xcolor}
\DeclareCaptionStyle{ruled}{labelfont=normalfont,labelsep=colon,strut=off}
\floatstyle{ruled}
\newfloat{listing}{tb}{lst}{}
\floatname{listing}{Listing}

\usepackage[acronym,nomain]{glossaries}
\makenoidxglossaries

\newacronym{ltl}{LTL}{linear temporal logic}
\newacronym{ltlf}{LTL$_f$}{linear temporal logic on finite traces}
\newacronym[shortplural=AP]{ap}{AP}{atomic proposition}
\newacronym{dfa}{DFA}{deterministic finite automaton}
\newacronym{fsm}{FSM}{finite state machine}
\newacronym{ast}{AST}{abstract syntax tree}
\newacronym{smt}{SMT}{satisfiability modulo theories}
\newacronym{pp}{pp}{percentage points}
\newacronym{ag}{A/G}{assume-guarantee}
\newacronym{ctl}{CTL}{computation tree logic}
\newacronym{pctl}{PCTL}{probabilistic computation tree logic}
\newacronym{dtmc}{DTMC}{discrete-time Markov chain}
\newacronym{dsl}{DSL}{domain-specific language}
\newacronym{pddl}{PDDL}{Planning Domain Definition Language}
\newacronym{pii}{PII}{personally identifiable information}

\newacronym{llm}{LLM}{large language model}
\newacronym{nl}{NL}{natural language}
\newacronym{mas}{MAS}{multi-agent system}
\newacronym{cps}{CPS}{cyber-physical system}
\newacronym{mcp}{MCP}{Model Context Protocol}
\newacronym{rag}{RAG}{retrieval-augmented generation}
\newacronym{sdk}{SDK}{software development kit}
\newacronym{api}{API}{application programming interface}

\newacronym{owasp}{OWASP}{Open Worldwide Application Security Project}
\newacronym{asi}{ASI}{Agentic Security Issue}
\newacronym{iam}{IAM}{identity and access management}
\newacronym{cve}{CVE}{common vulnerabilities and exposures}
\newacronym{sop}{SOP}{standard operating procedure}
\newacronym{rbac}{RBAC}{role-based access control}
\newacronym{otel}{OTEL}{OpenTelemetry}

\newacronym{kpi}{KPI}{key performance indicator}
\newacronym{asr}{ASR}{attack success rate}
\newacronym{fp}{FP}{false positive}
\newacronym{fn}{FN}{false negative}
\newacronym{fpr}{FPR}{false-positive rate}
\newacronym{tpr}{TPR}{true-positive rate}
\newacronym{sota}{SOTA}{state of the art}

\newtheorem{myproblem}{\textbf{Problem}}
\newtheorem{mydefinition}{\textbf{Definition}}

\newtheorem{myexample}{\textbf{Example}}

\newtheorem{myprop}{\textbf{Proposition}}

\newcommand{\yifeng}[1]{{\color{black}#1}}
\definecolor{nuzzopink}{RGB}{204,0,102}

\newcommand{\sysname}{${\mathsf{ContrAgent}}$\xspace}

\definecolor{cgaccent}{RGB}{214,110,28}
\definecolor{cgblue}{RGB}{43,108,176}

\graphicspath{{image/}}

\title{
    Symbolic Temporal Supervision of LLM Agents Using Contracts
}

\author{
    Yifeng Xiao\textsuperscript{\rm 1,2}\hspace{2.5em}
    Johnna Liu\textsuperscript{\rm 2}\hspace{2.5em}
    Pierluigi Nuzzo\textsuperscript{\rm 1}
}
\affiliations{
    \textsuperscript{\rm 1}Department of Electrical Engineering and Computer
    Sciences, University of California, Berkeley\\
    \textsuperscript{\rm 2}Sponsio Labs, Inc.\\
    \texttt{\{yifeng\_xiao, pierluigi.nuzzo\}@berkeley.edu}
}
\begin{document}

\maketitle

\begin{abstract}

Large language model (LLM) agents augmented by tools can automate complex, multi-step tasks, such as web navigation, code generation, and workflow orchestration, by acting on external systems through tool calls.
However, hallucinations, distributional instability, and adversarial manipulations in LLMs, and the irreversible consequences of certain tool calls can lead to harmful outcomes.
\yifeng{These risks call for safeguards that can check requirements across a trajectory, both before actions execute and when recorded behavior is evaluated.}
We present \sysname, a contract-based framework for symbolic temporal supervision of LLM agents.
\sysname captures an agent's behavior as a sequence of tool calls and formalizes it as a trace 
over a fixed set of checkable predicates. It then specifies required behaviors using assume-guarantee contracts in linear temporal logic over finite traces (LTL$_f$). Each contract is compiled to a deterministic finite automaton (DFA) that serves two roles: gating agent actions online and evaluating recorded traces offline.
A contract library, acting as a reusable knowledge base, is maintained independently of the agent's model and can be applied across different agents within the same task domain.
We show the effectiveness of our approach on four benchmarks spanning both roles, where \sysname~is competitive with LLM-judge, classifier, and guardrail baselines while producing deterministic, reproducible verdicts and, in the online mode, orders of magnitude lower per-call latency.
\end{abstract}

\section{Introduction}
\label{sec:intro}

Recent tool-augmented \gls{llm} agents {can} automate complex, multi-step tasks in real-world domains, including web navigation \cite{shum2023webagent}, code generation \cite{li2023code}, and workflow orchestration \cite{xu2023orchestrating}. 
Agentic frameworks such as OpenClaw \cite{openclaw}, HermesAgent \cite{hermesagent}, and LangChain \cite{langchain} achieve these capabilities through tool-call interfaces, ranging from shell commands to \gls{api} calls and protocols such as the \gls{mcp} \cite{mcp2024}.
These interfaces let agents act on external systems, but they also introduce safety risks: {the {hallucinations} and distributional instability of \gls{llm}s combined with possible adversarial manipulations and the irreversible consequences of tool calls can lead to harmful outcomes}.
These failure modes include indirect prompt injection, metric gaming under performance pressure, dangerous code execution, and identity confusion across delegating agents~\cite{john2025owasp}.

\yifeng{Several approaches grade recorded trajectories with an \gls{llm} judge or a learned detector~\cite{zheng2023judge, sharma2026willful, wen2025policyguard}. Runtime safeguards instead intervene during execution, using event-triggered rules~\cite{wang2025agentspec}, learned input-output checks~\cite{inan2023llamaguard}, verified policy code~\cite{miculicich2025veriguard}, temporal constraints during generation~\cite{kamath2025agentc}, or risk estimates from a learned agent model~\cite{wang2025pro2guard}. We seek a shared specification for the two supervision roles: an explicit temporal contract whose checker can gate proposed actions and evaluate recorded trajectories without a model in the checking loop.}

\begin{figure}[t!]
\centering
\includegraphics[width=0.95\columnwidth]{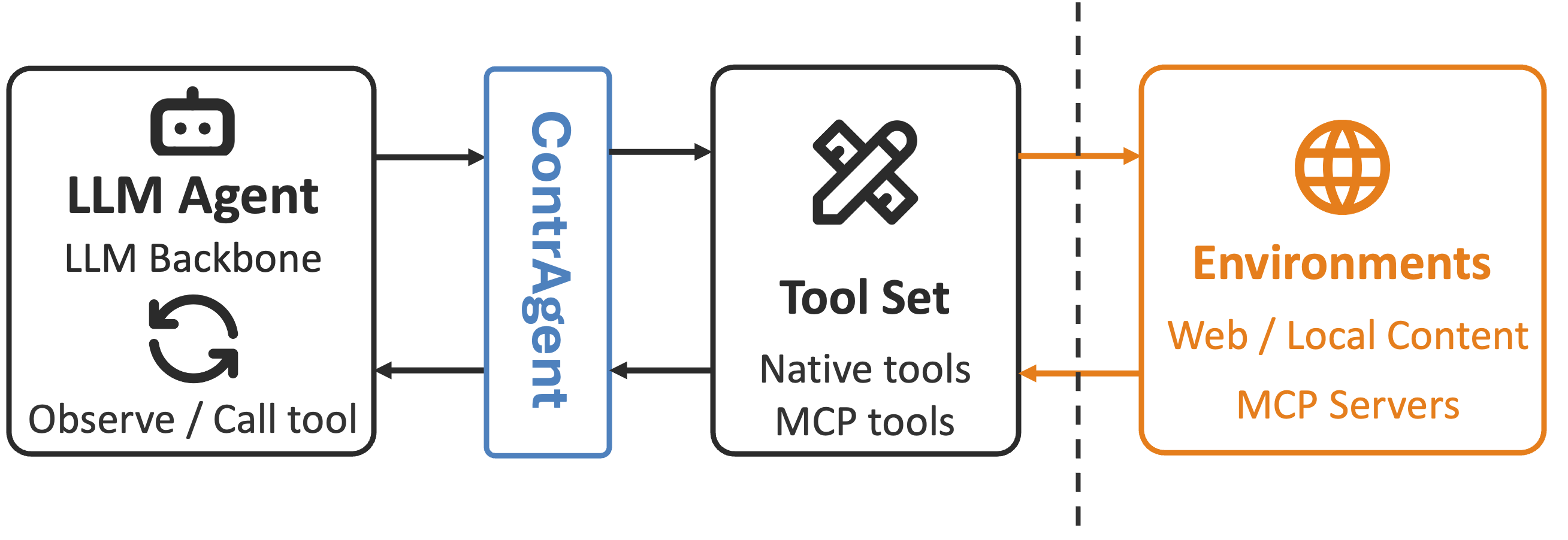}
\vspace{-2mm}
\caption{\sysname monitors the tool-call interface between an \gls{llm} agent
and its tools.}
\label{fig:agent}
\end{figure}

Inspired by \gls{ag} contract-based design {\cite{benveniste2018contracts,nuzzo2015methodology}}{, whose compositional reasoning has been applied to cyber-physical systems~\cite{xiao2024architecture,xiao2026architecture,xiao2026memocode} and to control synthesis with \gls{llm} agents~\cite{xiao2026acies}},
we present \sysname, a contract-based, model-agnostic, deterministic framework for symbolic temporal supervision of \gls{llm} agents.
\yifeng{For example, a refund policy can require approval before a refund and bound the refund amount. Such a policy depends on both the current call and earlier actions.}
We formalize trajectory-level requirements as \gls{ag} contracts in \gls{ltlf} \cite{degiacomo2013ltlf} over checkable \glspl{ap} on agent actions.
Contracts are compiled to \gls{dfa} checkers that supervise the agent's trajectory, by gating agent actions online, as shown in Figure~\ref{fig:agent}, or evaluating recorded traces offline.
\yifeng{We call these propositions \emph{interaction predicates}: deterministic tests of tool calls and results, model input and output, context, permissions, and counters. Temporal operators combine these tests into requirements on the trajectory. An assumption states the environment conditions under which the agent must provide the guarantee. Separating environment predicates from agent predicates supports contract analysis and distinguishes assumption failures from guarantee violations. \sysname checks individual contracts for consistency and compatibility and analyzes the library for conflicts. This formulation is also a first step toward the composition and refinement of agent contracts.}
Our contributions can be summarized as follows:

\begin{itemize}\setlength{\itemsep}{2pt}

\item \yifeng{We introduce \gls{ag} contracts for interface-observable behavioral requirements of \gls{llm} agents over a fixed \mbox{predicate vocabulary}.}

\item We present \sysname, a model-agnostic, deterministic framework that uses the same contracts to shield an agent online and score its traces offline.

\item We evaluate \sysname on four benchmarks and show that it is competitive with LLM-judge, classifier, and guardrail baselines, at a per-call latency that is orders of magnitude lower.

\end{itemize}

The remainder of the paper is organized as follows. After discussing related work in Section~\ref{sec:related}, we provide background on agent frameworks, \gls{ltlf}, and \gls{ag} contracts in Section~\ref{sec:background}. Section~\ref{sec:contracts} introduces interaction predicates and agent contracts, while Section~\ref{sec:framework} provides details on contract compilation and contract-based supervision. Section~\ref{sec:experiments} reports results from the four benchmarks, and is followed by concluding remarks in Section~\ref{sec:conclusion}.

\section{Related Work}
\label{sec:related}

{Most runtime safeguards for tool-based operation of \gls{llm} agents act at the tool-call boundary. \yifeng{AgentSpec~\cite{wang2025agentspec} combines event triggers, predicates over the current execution context, and enforcement actions. Our formulation expresses trajectory requirements directly as temporal A/G contracts, separating environment assumptions from agent guarantees and analyzing the resulting contract library.}
A separate information-flow line of work tracks data and privilege across the execution to stop prompt injection and exfiltration, as in Progent \cite{shi2025progent}, CaMeL \cite{debenedetti2025camel}, RTBAS \cite{zhong2025rtbas}, and Fides \cite{costa2025ifc}. Task Shield \cite{jia2025taskshield} checks that each tool call serves the user's goal, and MELON \cite{zhu2025melon} detects injections by re-executing the trajectory with a masked user prompt. \yifeng{These defenses track information flow or task alignment; our contracts specify observable behavioral requirements.}

A class of approaches uses logic to reason about the execution trace: {Agent-C \cite{kamath2025agentc} checks temporal constraints with \gls{smt} solving during generation, VeriGuard \cite{miculicich2025veriguard} generates policy code with an \gls{llm} and formally verifies it before use, and FORGE \cite{palumbo2026forge} weaves deterministic Datalog policies into multi-agent deployments. These methods differ from ours in what their policies can express and in how they reach a verdict. FORGE's Datalog policies have no temporal operators, so a policy can require a past condition but not a follow-up action, and each check returns permit or deny. Agent-C solves an \gls{smt} query over the trace at each call and relies on token-level control of the model. Neither FORGE nor Agent-C separates environment assumptions from agent guarantees. Guardrail gateways such as Invariant Guardrails \cite{invariant2025guardrails} match rule patterns over the trace, e.g., one call followed by another, without a formal semantics or a check for conflicting rules. Safety Chip \cite{yang2024safetychip} compiles propositional \gls{ltl} to mask unsafe actions of embodied robot agents, whereas \sysname supervises tool-calling \gls{llm} agents with \gls{ag} contracts over checkable \glspl{ap}.}}

{Another class of approaches aims to grade or shield behaviors using a
learned model or an \gls{llm} judge. \gls{llm}-as-judge graders
\cite{zheng2023judge}, AgentPex
\cite{sharma2026willful}, and learned detectors such as PolicyGuard
\cite{wen2025policyguard} score recorded trajectories, but their verdicts are
stochastic and hard to reproduce, particularly on safety judgments \cite{chen2025saferluckier}.
ShieldAgent \cite{chen2025shieldagent} compiles policy documents into action-based
probabilistic rule circuits, ABC \cite{bhardwaj2026abc} enforces behavioral
contracts under a probabilistic satisfaction relation, and \yifeng{ProbGuard~\cite{wang2025pro2guard} estimates future risk using a learned agent model and triggers interventions at a risk threshold. These approaches use probabilistic assessments; \sysname instead checks logical requirements over deterministically grounded action traces.}}

\section{Preliminaries}
\label{sec:background}

{As we aim to supervise an agent's tool calls with \gls{ag} contracts, we first give background on agent frameworks and temporal logic, then state the supervision problem.}

\subsection{Agent Frameworks}

An agent pairs an \gls{llm} backbone with a tool set, issuing tool calls and folding their outputs into a reasoning loop until task completion~\cite{yao2023react}.
{The \emph{tool-call interface} is the set of tools the framework exposes to the agent, together with their call and return conventions.}
{Reasoning steps carry the agent's decision making, but they take
external effect only through the tool calls they issue, and the class of
irreversible consequences this paper targets arises at this interface.}
\sysname therefore tracks the \emph{tool-call trace}, the ordered
sequence of tool-call events the agent emits, now often a {long-horizon,}
multi-tool trajectory~\cite{xu2026tooluse}.

\begin{mydefinition}[{Agent session}]
\label{def:session}
\yifeng{Let $\mathcal{E}$ be the set of interface events, including agent-issued tool calls and model outputs, tool returns, and model inputs. An execution is a transition system~\cite{baierkatoen2008} $\mathcal{T}=(S,s_0,\mathcal{E},R)$, with $R\subseteq S\times\mathcal{E}\times S$. A state $s=(\eta,\kappa,\rho)$ contains the event history $\eta\in\mathcal{E}^{*}$, a finite map $\kappa$ of context keys to values, and a vector $\rho$ of session counters; initially $s_0=(\varepsilon,\kappa_0,\rho_0)$. A transition $(s,b,s')\in R$ appends event $b$ to $\eta$ and updates $\kappa$ and $\rho$. A session is a finite path $s_0\xrightarrow{b_1}\cdots\xrightarrow{b_m}s_m$, with event history $\eta=b_1\cdots b_m$.}

\yifeng{A fixed deterministic abstraction $\pi:\mathcal{E}^{*}\to\Sigma^{*}$ maps the event history to the monitored action trace $\tau=\pi(\eta)=a_1\cdots a_k$. Each agent action contributes one position; $\Sigma$ is the set of action records containing its call or output fields and associated observations. Tool returns enrich the corresponding call record, while inputs update the available context; neither introduces an additional action position. The result field remains unavailable until the tool returns. Thus, $\eta$ retains all interface events, whereas temporal operators over $\tau$ count agent actions: $X$ refers to the next action position. Both alphabets may be infinite because event and record fields range over unbounded values.}
\end{mydefinition}
{A context key records a fact carried across events, such as the caller's identity or a document's source, and a counter tracks a cumulative quantity, such as the tokens consumed.}

\subsection{Finite-Trace LTL (LTL$_f$)}
{\gls{ltl}~\cite{pnueli1977temporal} extends propositional
logic to reason over infinite sequences of propositional
interpretations, or \emph{traces}.} \gls{ltlf}~\cite{degiacomo2013ltlf}
{restricts this logic to finite traces. Let $AP$ be a
finite set of atomic propositions. The class of \gls{ltlf} formulae over $AP$ is
defined by the grammar}
$
  \varphi ::= P \mid \neg\varphi \mid \varphi \wedge \varphi'
  \mid X\,\varphi \mid F\,\varphi \mid G\,\varphi \mid \varphi\,U\,\varphi',
$
where $P \in AP$ is an \gls{ap} and $\varphi'$ is an \gls{ltlf}
formula, with temporal operators $X$ (next), $F$ (eventually), $G$ (always), and
$U$ (until).
{Each \gls{ltlf} formula compiles to an equivalent \gls{dfa} that accepts exactly the traces satisfying it~\cite{degiacomo2013ltlf}, which is the canonical monitor construction of runtime verification~\cite{bauer2011rvltl,leucker2009brief}.}

{When an atomic predicate is used to compare an arithmetic quantity read from
the session state $s$, such as a call count or token total accumulated in $\rho$, or an argument value latched into $\kappa$, against a bound, then it is better captured by a linear-arithmetic constraint than by a plain proposition. An \gls{ltlf} formula over such atoms is then taken modulo linear arithmetic, i.e., arithmetic LTL$_f$ (ALTL$_f$)~\cite{felli2023monitoring}, an instance of \gls{ltlf} modulo theories~\cite{geatti2022ltlfmt}. \sysname evaluates every atom \emph{pointwise}, i.e., its truth value at each event is computed from the current event $a$ and the session state $s$.} Because every atom is decided pointwise, the monitored logic is propositional \gls{ltlf} over the finite alphabet of atom valuations. We keep the ALTL$_f$ notation for readability and do not rely on the lookahead results of \citet{felli2023monitoring}, which address atoms whose value depends on future events.

\subsection{Assume-Guarantee (A/G) Contracts}

Let $M$ denote a component, i.e., an element of a system, characterized by a set of variables $V_c$ and a set of behaviors $\llbracket M \rrbracket$ over $V_c$.
A contract $C$ formally captures a set of specifications for $M$ using a triple $C = (V_c, A, G)$~\cite{benveniste2018contracts}, where $A$ and $G$ are sets of behaviors over $V_c$. $A$ is the assumptions on the environment of $M$ while $G$ is the guarantees provided by $M$, given that the assumptions are satisfied. 
We say that $M$ is a valid implementation of $C$, i.e., $M \models C$, if all the behaviors of $M$ are included in the guarantees given the assumptions of $C$, i.e., $\llbracket M \rrbracket \subseteq G \cup \overline{A}$. We say that component $E_c$ is a valid environment of $C$ if all the behaviors of $E_c$ are contained in
the assumptions of $C$.
\yifeng{For example, an assumption may require a policy service to provide valid authorization information, while the guarantee requires the agent to issue refunds only within the authorized limit. The assumption describes what the environment must supply; the guarantee describes what the agent must do under that condition. A trigger inside the guarantee, such as a refund request, instead creates an obligation during the run.}
A contract is \emph{consistent} if and only if there exists a valid implementation,
i.e., $G \cup \overline{A} \ne \emptyset$, and it is \emph{compatible} if there
exists a valid environment $E_c$, i.e., $A \ne \emptyset$.

{\sysname formalizes contracts over the agent's tool-call
trace: {the component is the agent, whose behaviors
$\llbracket M \rrbracket$ are the tool-call traces it can produce
(Def.~\ref{def:session})}, and the variables $V_c$ are the set $AP$.
{At runtime, a session exposes a single behavior of this component, and} \sysname evaluates {the observed} trace $\tau$ by 
the three-valued valuation below.}

\begin{mydefinition}[{Runtime contract valuation}]\label{def:contract}
A contract is a triple $C = (V, A, G)$, the instance of $(V_c, A, G)$ in which $V$ is a set of atomic propositions and the assumption $A$ and the guarantee $G$ are \gls{ltlf} formulas over $V$, each standing for the set of traces that satisfy it. A prefix can violate a formula and still have an extension that satisfies it. For a prefix $u$ and a formula $\varphi$, we therefore use the four-valued verdict of runtime verification \cite{bauer2010comparing,degiacomo2022monitoring}: $[u \models \varphi] = \top_p$ (resp.\ $\bot_p$) if every finite extension $uv$ satisfies (resp.\ violates) $\varphi$, and otherwise $[u \models \varphi] = \top_c$ (resp.\ $\bot_c$) if $u$ itself satisfies (resp.\ violates) $\varphi$.
On a prefix $u$, the valuation of $C$ takes one of three values, named {$\mathrm{IDLE}$ ($1$), $\mathrm{ACTIVE}$ ($e$), and $\mathrm{FAIL}$ ($0$)}, ordered as $0 \le e \le 1$, following the multi-valued verdicts of runtime verification~{\cite{bauer2006monitoring}}:
\begin{equation}\label{eq:valuation}
\llbracket C\rrbracket(u) =
\begin{cases}
1, & [u \models A] = \bot_p,\\
0, & [u \models A] \neq \bot_p \text{ and } [u \models G] = \bot_p,\\
e, & \text{otherwise.}
\end{cases}
\end{equation}
When the session ends, the trace $\tau$ is complete and the current verdicts become final. The end valuation $\llbracket C\rrbracket^{\mathrm{end}}(\tau)$ is $1$ if $\tau \not\models A$, $0$ if $\tau \models A \wedge \neg G$, and $e$ otherwise.
\end{mydefinition}

\yifeng{$\mathrm{ACTIVE}$ means that neither an assumption failure nor a guarantee violation has been established permanently; an obligation may still be pending. It does not mean that every required action has already occurred. For example, take $A=\top$ and the guarantee $G(\mathit{trig} \rightarrow F\,\mathit{resp})$, where $\mathit{trig}$ is a trigger and $\mathit{resp}$ its required response. After a trigger without a response, $[u \models G]=\bot_c$ and the contract remains $\mathrm{ACTIVE}$ because a later response can still discharge the obligation. If the session ends without that response, its end valuation is $\mathrm{FAIL}$.}

\subsection{Agent Trajectory Supervision}
\label{sec:problem}

{An execution monitor can be used to halt a run at the first policy-violating action, thus enforcing a safety
property~\cite{schneider2000enforceable}. We adapt this approach to the agent's tool-call trace.}

\begin{myproblem}[Agent Trajectory Supervision]\label{def:supervision}
Agent trajectory supervision is the problem of monitoring an agent's tool-call trace $\tau$ against a set of contracts $\mathcal{C} = \{C_1, \dots, C_n\}$ with $C_i = (V_i, A_i, G_i)$. The monitor verdict $v(\tau) = \big(v_{\mathrm{env}}(\tau),\; v_{\mathrm{ag}}(\tau)\big)$ is defined by the join and the meet of the contract valuations as follows:
\begin{align}
v_{\mathrm{env}}(\tau) = \textstyle\bigvee_{i} \llbracket C_i \rrbracket(\tau), \\
v_{\mathrm{ag}}(\tau) = \textstyle\bigwedge_{i} \llbracket C_i \rrbracket(\tau).
\end{align}
The environment is not valid for at least one contract if and only if $v_{\mathrm{env}}(\tau) = 1$; the agent violates at least one contract if and only if $v_{\mathrm{ag}}(\tau) = 0$. Therefore, {trajectory supervision} provides
deterministic maps over $\mathcal{C}$ as follows:
\begin{itemize}\setlength{\itemsep}{1pt}
\item In \emph{online enforcement}, it provides a map
$\sigma : \Sigma^{*} \to \{\mathsf{pass}, \mathsf{block}\}$ that, on the trace, returns $\mathsf{block}$ at the first prefix
$\tau_k = a_1 \cdots a_k$ with {$v(\tau_k) \ne (e, e)$} and $\mathsf{pass}$ at every earlier prefix, so an event is rejected only if it permanently violates an assumption or a guarantee;
\item In \emph{offline evaluation}, it provides a map
{$\mu : \Sigma^{*} \to \{0, e, 1\}^{2}$} with $\mu(\tau) = v^{\mathrm{end}}(\tau)$, the monitor verdict computed from the end valuations of Def.~\ref{def:contract}.
\end{itemize}
\end{myproblem}

\begin{myprop}[Online/offline agreement]\label{prop:agree}
Let $\tau$ be a complete trace and $C = (V, A, G)$ a contract with $\tau \models A$. If $G$ is a safety property, i.e., every trace that violates $G$ has a prefix $u$ with $[u \models G] = \bot_p$, then $\llbracket C \rrbracket^{\mathrm{end}}(\tau) = 0$ if and only if $\llbracket C \rrbracket(u) = 0$ for some prefix $u$ of $\tau$.
\end{myprop}
\noindent\emph{Proof sketch.} If $\tau \not\models G$, the safety of $G$ gives a prefix $u$ with $[u \models G] = \bot_p$, while $[u \models A] \neq \bot_p$ since $\tau \models A$. Conversely, $[u \models G] = \bot_p$ implies $\tau \not\models G$. \hfill$\diamond$

\sysname solves this problem using one checker per contract, executed online for $\sigma$ and replayed offline for $\mu$, as shown in \S\ref{sec:framework-modular}.

\section{Agent Contracts}
\label{sec:contracts}

We introduce the \textit{interaction predicate}, the atomic proposition from which agent contracts are built.
In the following, $\top$ and $\bot$ denote the Boolean values \emph{true} and \emph{false}.

\begin{mydefinition}[Interaction Predicate]\label{def:ap}
We define an \emph{interaction predicate} as a predicate $P(s, a, c)$ over a session state $s$, \yifeng{an action record $a \in \Sigma$} (Def.~\ref{def:session}), and a parameter $c$. Its truth value, $\top$ or $\bot$, is a deterministic function of $s$, $a$, and $c$, evaluated by the monitor at each event (\S\ref{sec:framework-compile}).
\end{mydefinition}

The parameter $c$ instantiates a predicate with, for example, a numeric bound or a string pattern.
Given a session, we define a set of interaction predicates monitored at runtime.
\yifeng{Predicates over a call and the session state are evaluated before execution. Result predicates, such as $\mathsf{OutHas}$, are evaluated when the result becomes available and the corresponding action record is updated (Def.~\ref{def:session}).}
We categorize the predicates into two families, as shown in Table~\ref{tab:atoms}.

\begin{table}[t]
\centering
\small
\setlength{\tabcolsep}{3pt}
\renewcommand{\arraystretch}{1.05}
\begin{tabularx}{\columnwidth}{@{}c l X@{}}
\toprule
\textbf{Type} & \textbf{Predicate} & \textbf{Meaning} \\
\midrule
\multirow{8}{*}{\rotatebox[origin=c]{90}{{Structural}}}
  & $\mathsf{Call}(T)$ & tool $T$ is invoked \\
  & $\mathsf{ArgHas}(T,f,p)$ & argument $f$ of $T$ matches pattern $p$ \\
  & $\mathsf{OutHas}(T,p)$ & result of $T$ matches pattern $p$ \\
  & $\mathsf{Said}(p)$, $\mathsf{In}(p)$ & model output / input matches pattern $p$ \\
  & $\mathsf{Match}(f,k)$ & field $f$ equals the value of context key $k$ \\
  & $\mathsf{Subset}(f,S)$ & values in field $f$ lie within set $S$ \\
  & $\mathsf{Flow}(g,g')$ & agent $g'$ reads data written by agent $g$ \\
  & $\mathsf{Perm}(Q)$ & caller holds permission $Q$ \\
\midrule
\multirow{3}{*}{\rotatebox[origin=c]{90}{{Numeric}}}
  & $\mathsf{Cnt}(T)$ & current number of $T$ calls \\
  & $\mathsf{Num}(T,f)$ & numeric value of argument field $f$ \\
  & $\mathsf{Tok}$ & cumulative tokens consumed \\
\bottomrule
\end{tabularx}
\caption{\sysname's interaction predicates (subset; full vocabulary in Appendix~\ref{app:atoms}). $T$ is a tool, $f$ an argument field, $p$ a regular-expression or literal-value pattern, $Q$ a permission set, $S$ an allowed-value set, $g,g'$ agents, and $k$ a context key. Numeric rows list the quantity $\theta$ of the predicate $\theta \bowtie c$.}
\label{tab:atoms}
\end{table}

\paragraph{Structural predicates.}
A \emph{structural} predicate is a deterministic Boolean test of a
discrete condition: the presence or absence of an event (e.g., a tool firing, $\mathsf{Call}$, or a data flow, $\mathsf{Flow}$), or a pattern, equality, or membership match on a field (e.g., $\mathsf{ArgHas}$, $\mathsf{Match}$), with no model in the loop.

\paragraph{Numeric predicates.}
We extend the structural family with a \emph{stateful, numeric} family that tracks cumulative quantities over the session, such as counts and totals.
A \emph{numeric} predicate has the form $\theta(s, a) \bowtie c$, with $\bowtie \in \{\le, <, \ge, >, =\}$, where $\theta$ extracts a quantity from the current event $a$ and state $s$, such as a count, a total, or an argument's value, and $c$ is a constant bound.
The predicate holds exactly when the comparison holds, $P(s,a,c) = \big[\,\theta(s, a) \bowtie c\,\big]$, where $[\cdot]$ is the Iverson bracket, as in $\mathsf{Cnt}(T) \le N$.
{The monitor maintains each accumulated quantity in a \emph{counter}, one entry of the vector $\rho$ of Def.~\ref{def:session}, updated deterministically at each event.}

With the \gls{ltlf} operators, we use interaction predicates to define agent contracts for trajectory supervision as follows.

\begin{mydefinition}[Agent Contract]\label{def:agentcontract}
An \emph{agent
contract} is a triple $C = (V, \varphi_A, \varphi_G)$, where:
\begin{enumerate}\setlength{\itemsep}{3pt}
\item {$V = V_{\mathrm{ag}} \cup V_{\mathrm{env}}$ is a finite set of interaction predicates (Def.~\ref{def:ap}), the \glspl{ap} of the contract's formulas, where $V_{\mathrm{ag}}$ collects the predicates decided by the agent's own actions and $V_{\mathrm{env}}$ those decided by its environment.}
\item {The {assumption} $\varphi_A$ is an ALTL$_f$ formula over $V_{\mathrm{env}}$, and the {guarantee} $\varphi_G$ is an ALTL$_f$ formula over $V$, both according to the grammar of \S\ref{sec:background}.}
\end{enumerate}
\end{mydefinition}

The partition of $V$ mirrors the distinction between controlled and uncontrolled variables in contract-based design~\cite{benveniste2008multiple}. {For example, $\mathsf{Call}$, $\mathsf{ArgHas}$, and $\mathsf{Cnt}$ track the tool calls that the agent itself issues (Def.~\ref{def:session}), their arguments, and the counts derived from them, so they belong to $V_{\mathrm{ag}}$. $\mathsf{OutHas}$, $\mathsf{In}$, and $\mathsf{Perm}$ track tool results, user input, and granted permissions, so they belong to $V_{\mathrm{env}}$. The actions of other agents also belong to the environment.}
Semantically, an agent's session satisfies the contract when its trace satisfies $\varphi_A \rightarrow \varphi_G$, so the agent contract is the triple $(V, A, G)$ of Def.~\ref{def:contract} with $A = \varphi_A$ and $G = \varphi_G$. Such contracts capture a wide range of trajectory properties over the tool-call interface, including order-, history-, and count-dependent ones (Appendix~\ref{app:temporal}).

\begin{figure}[t]
\centering
\begin{tikzpicture}[font=\scriptsize,
  box/.style={rounded corners=3pt, draw, line width=0.8pt, align=left, inner sep=3pt},
  hd/.style={font=\footnotesize\bfseries}]
\node[box, draw=cgblue, fill=cgblue!5, text width=0.86\columnwidth, align=center] (ltl) {%
  {\bfseries\footnotesize \textcolor{cgblue}{Temporal Logic (ALTL$_f$)}}\\[-1pt]
  {operators $G\ \, F\ \, X\ \, U$ lift predicates over the trajectory}\\[-1pt]
  \textcolor{black!55}{$\lnot\mathsf{Call}(\mathsf{refund})\,U\,\mathsf{Call}(\mathsf{approve})\,\wedge\,G\big(\mathsf{Call}(\mathsf{refund})\!\to\!\mathsf{Num}(\mathsf{refund},\mathsf{amount}) \,\le\, 42\big)$}};
\node[box, draw=black!60, fill=black!3, text width=0.40\columnwidth, align=center, anchor=north west]
  at ([yshift=-4mm]ltl.south west) (cat) {%
  {\bfseries\footnotesize Structural Predicates}\\[-1.5pt]
  {presence or pattern match}\\[-1pt]
  \textcolor{black!55}{$\mathsf{Call}(\mathsf{refund})=\top$}\\[-1.5pt]
  \textcolor{black!55}{$\mathsf{Match}(\mathsf{account},\mathsf{appr})=\top$}\\[-1.5pt]
  \textcolor{black!55}{$\mathsf{Call}(\mathsf{approve})=\top$}};
\node[box, draw=cgaccent, fill=cgaccent!7, text width=0.40\columnwidth, align=center, anchor=north east]
  at ([yshift=-4mm]ltl.south east) (num) {%
  {\bfseries\footnotesize \textcolor{cgaccent}{Numeric Predicates}}\\[-1.5pt]
  {threshold over accumulators}\\[-1pt]
  \textcolor{black!55}{$\mathsf{Cnt}(\mathsf{refund})=3\le 5$}\\[-1.5pt]
  \textcolor{black!55}{$\mathsf{Tok}=1200\le 2000$}\\[-1.5pt]
  \textcolor{red}{$\mathsf{amount}=50>42$}};
\node[box, draw=black!45, fill=white, text width=0.86\columnwidth, align=center, anchor=north west]
  at ([yshift=-4mm]cat.south west) (ev) {%
  {tool-call event}\ \ \textcolor{black!55}{$a=(\mathsf{refund},\ \{\mathsf{amount}{:}\,50,\dots\})$}};
\draw[-{Stealth[length=3pt]}, line width=0.5pt] (ev.north -| cat.south) -- (cat.south);
\draw[-{Stealth[length=3pt]}, line width=0.5pt] (ev.north -| num.south) -- (num.south);
\node[above=0.2mm of ev, font=\scriptsize] {extract / accumulate};
\draw[-{Stealth[length=3pt]}, line width=0.5pt] (cat.north) -- (cat.north|-ltl.south);
\draw[-{Stealth[length=3pt]}, line width=0.5pt] (num.north) -- (num.north|-ltl.south);
\node[font=\scriptsize] at ([yshift=-2mm]ltl.south) {Interaction Predicates $V$};
\end{tikzpicture}
\caption{{One tool-call event grounded into structural and numeric
predicates, then lifted by the ALTL$_f$ layer.}}
\label{fig:twolayer}
\end{figure}
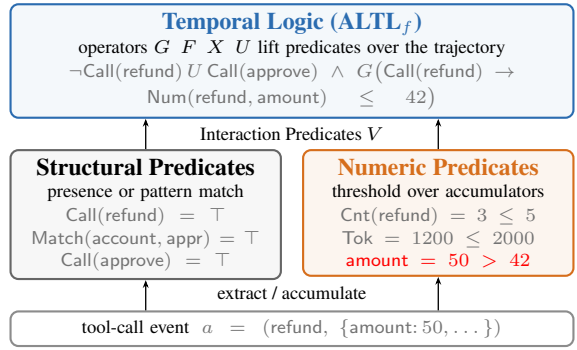

\begin{figure*}[!t]
\centering
\includegraphics[width=\textwidth]{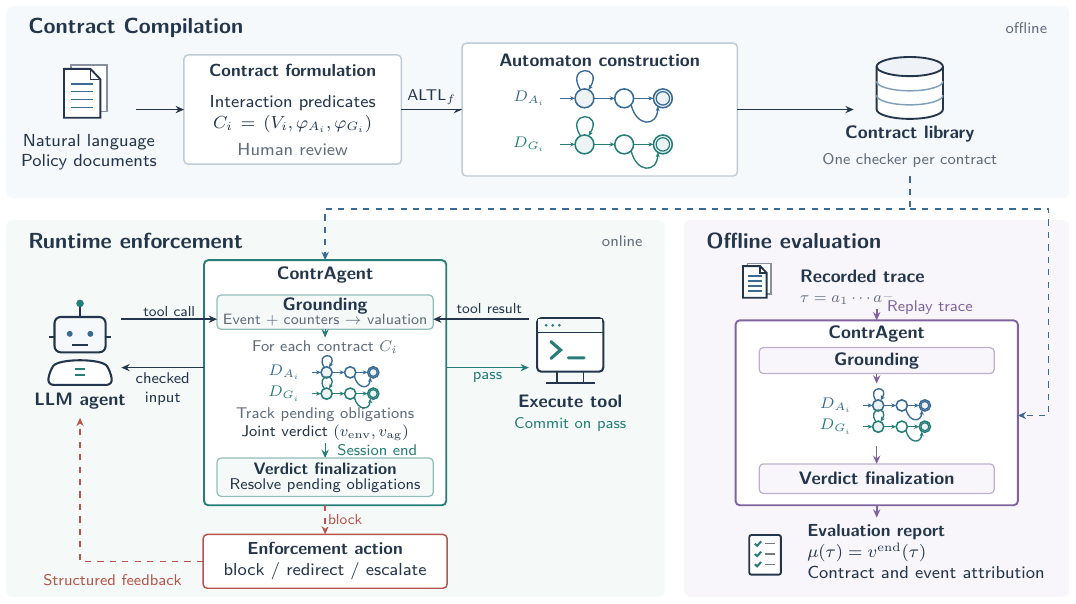}
\caption{The \sysname framework: authoring inputs compile once into a
contract checker (top), read online to block the first violating call
(bottom left) and offline to score a recorded trace (bottom right).}
\label{fig:overview}
\end{figure*}

\begin{myexample}\label{ex:grounding}
\yifeng{Consider the guarantee in Figure~\ref{fig:twolayer} with assumption $\varphi_A=\top$. This instance requires an approval to occur, forbids refunds before that approval, and fixes the refund limit at \$42. Suppose approval has occurred and the agent proposes a refund of \$50. The call predicate $\mathsf{Call}(\mathsf{refund})$ holds, but $\mathsf{Num}(\mathsf{refund},\mathsf{amount})\le 42$ is false. The guarantee therefore fails permanently, and the contract falls to $\mathrm{FAIL}$ before the refund executes. The account-matching predicate illustrated in the figure is available for additional requirements but is not part of this guarantee.}
\end{myexample}

\section{The \sysname Framework}
\label{sec:framework}

\yifeng{\sysname first prepares a contract library, then uses it for supervision. During \emph{contract preparation} (\S\ref{sec:framework-compile}), requirement artifacts are expressed as agent contracts and translated into deterministic monitors. During \emph{supervision} (\S\ref{sec:framework-modular}), the same monitors either gate proposed actions online or evaluate recorded actions offline. Preparation takes place before either use; offline evaluation refers to replaying a completed~trace.}

\subsection{Contract Compilation}
\label{sec:framework-compile}

Contract compilation extracts contracts from requirement artifacts (e.g., natural language, policy documents, recorded traces) into ALTL$_f$ formulas over the interaction predicates of \S\ref{sec:contracts}, then compiles them into \glspl{dfa} (Figure~\ref{fig:overview}). {The two steps are described next.}

\paragraph{Contract formulation.}
\label{sec:framework-authoring}
{Besides manually written ALTL$_f$ contracts, we support translating natural-language requirements and extracting rules from policy documents.}
\yifeng{Following a lift-then-ground decomposition~\cite{lang2ltl}, the \gls{llm} first \emph{lifts} each utterance to an ALTL$_f$ formula with placeholder leaves. A deterministic \emph{predicate-binding} step then maps the placeholders to interaction predicates from Table~\ref{tab:atoms}, yielding the closed formula. This authoring step selects the predicates; runtime grounding evaluates their truth values on actual actions.}
The \gls{llm} thus fills templates over the fixed vocabulary of Table~\ref{tab:atoms} ~\cite{wang2023grammar} with a few prompt examples. Each formula is translated back to natural language for human review~\cite{cosler2023nl2spec}.

\paragraph{Automaton construction.}

By the standard \gls{ltlf}-to-\gls{dfa} construction~\cite{degiacomo2013ltlf},
every contract formula over $V$ compiles to a \gls{dfa} whose alphabet is
the set $2^{V}$ of predicate valuations. The \gls{dfa} reads a trace one
event at a time, taking a transition on the predicate valuation at that event, and accepts exactly the finite traces
that satisfy the formula.
{{We call \emph{grounding} the deterministic step that evaluates each predicate on the current event and the counters, producing the valuation read by the \gls{dfa}.} For a numeric predicate, grounding evaluates the arithmetic
constraint $\theta(s,a)\bowtie c$ against the session's counters,
and the \gls{dfa} construction stays standard. The counter state is read {only during grounding} and is not encoded in the \gls{dfa}, which keeps it
finite.}
A contract carries two such formulas, the assumption $\varphi_A$ and the guarantee $\varphi_G$, so each compiles to its own \gls{dfa}.
We store the two \glspl{dfa} as one checker corresponding to the contract for agent supervision.

\paragraph{Contract library.}
\label{sec:patterns}
The compiled checkers form a \emph{contract library} that \sysname loads at
runtime. Because a contract is written over interaction predicates rather than any model's internals, the library is portable: the same checkers apply unchanged across agents and \gls{llm} models that share a tool interface, so a domain's policies transfer between models. 
\yifeng{\sysname checks each contract for consistency and compatibility. To analyze interactions within the library, it tests the conjunction $\bigwedge_i(A_i\wedge G_i)$, which asks whether all contracts can apply and have their guarantees satisfied together. An unsatisfiable conjunction requires further analysis to distinguish incompatible assumptions from conflicting guarantees.} \yifeng{We treat comparison atoms as propositions and add} the arithmetic relations between them as background conjuncts, e.g., $\mathsf{Cnt}(T) \le 5$ excludes $\mathsf{Cnt}(T) > 7$. When the conjunction is unsatisfiable, we enumerate all minimal unsatisfiable cores~\cite{roveri2024ltlfuc,ielo2026ltlfmuc} and report a conflict for every core whose assumptions are jointly satisfiable. A core whose assumptions are already unsatisfiable marks contracts that never apply together, which is not a conflict.

\subsection{Contract-Based Supervision}
\label{sec:framework-modular}

At each event, \sysname evaluates every interaction predicate and advances each loaded contract's two \glspl{dfa} on the resulting valuation. {The two runs decide the contract's value in Def.~\ref{def:contract}, and the values of all loaded contracts give the verdict $v(\tau)$ of Problem~\ref{def:supervision}.}

\yifeng{Each contract $(V,A,G)$ is checked independently, so a report identifies the contract, the action involved, and whether the assumption or guarantee failed. An operator can inspect the corresponding trace to diagnose the agent behavior, environment condition, or policy specification. When assumptions depend on future observations, attribution remains conditional until the completed trace is evaluated. This modularity also lets the system enforce multiple policies at once, without merging them into a monolithic formula.}

\paragraph{Runtime enforcement.}
At runtime, \sysname realizes the enforcement map $\sigma$ of Problem~\ref{def:supervision}. 
\yifeng{As shown in Figure~\ref{fig:overview}, the monitor checks a proposed call before the tool executes. It evaluates the candidate action and its counter updates, and rejects the call if a contract reports a violation; a rejected candidate is rolled back from the monitored trace. The response to an allowed call enriches its action record and is checked before delivery to the agent, without adding another action step. Incoming inputs or results that violate an assumption can be \emph{suppressed}, filtering the observations delivered at the interface, as in bidirectional runtime enforcement~\cite{aceto2021bidirectional}. Suppressing a result does not undo effects of the tool execution that produced it.}
The check is incremental. \sysname does not construct the two \glspl{dfa} of a contract in full. For the assumption and for the guarantee, it keeps a residual formula and progresses it one event at a time~\cite{rosu2005}. Each residual plays the role of a \gls{dfa} state, so the two \glspl{dfa} are advanced on the fly, and the per-event cost depends on the contract formulas and on grounding, but not on the trace length. {Supervision thus scales to long horizons.} Progression reports a permanent violation when the residual simplifies to false. This report is sound, but it can come later than the first prefix $u$ with $[u \models \varphi] = \bot_p$.

{When the blocked event is a call, the} enforcement action is decided by the strategy carried by the violated contract: \emph{block} the call (the default), \emph{redirect} it to a safe alternative, or \emph{escalate} to a human. 
The outcome returns to the model as a structured message: the
identifier of the {contract involved}, a short explanation, and, where it
applies, the suggested replacement or required next action, so the agent can steer its trajectory back into the safe region. For example, a contract that forbids issuing a refund before approval can return a message like ``the action \texttt{issue\_refund} was rejected: call \texttt{check\_policy} first''.

\yifeng{\sysname also prompts the agent to discharge an obligation as soon as its trigger occurs. For a prescriptive guarantee $G(\mathit{trig} \rightarrow X\,\mathit{tool})$, an accepted trigger produces feedback asking the agent to call the required tool on its next action. Here $X$ denotes the next agent action, not the return of the current tool call (Def.~\ref{def:session}). If the agent proposes a different action, the monitor rejects it. If the session ends before the required action occurs, the obligation is unsatisfied at termination. Thus, the feedback guides the agent, while the contract checks whether it follows the instruction.}
An unbounded liveness guarantee $G(\mathit{trig} \rightarrow F\,\mathit{resp})$ cannot be refuted by any finite prefix: after $\mathit{trig}$ is satisfied, it stays pending until $\mathit{resp}$ occurs. For example, a guarantee that requires a confirmation after each refund stays pending until the agent sends it. At the end of the agent session, \sysname surfaces every still-pending guarantee as a feedback message asking the agent to perform $\mathit{resp}$, and a guarantee left unsatisfied collapses to $\mathrm{FAIL}$, as given by the end valuation of Def.~\ref{def:contract}. \yifeng{Proposition~\ref{prop:agree} relates the exact prefix and end verdicts for safety guarantees. Unbounded response obligations additionally require an end-of-session check, since an unfinished prefix may still be extended with the required response.}

\paragraph{Offline evaluation.}
Contracts also serve as a deterministic evaluation engine for recorded traces: {$\mu : \Sigma^{*} \to \{0, e, 1\}^{2}$} of Problem~\ref{def:supervision}, as shown in Figure~\ref{fig:overview}.
\sysname replays a recorded trace and returns its end-of-trace verdict
$\mu(\tau) = v^{\mathrm{end}}(\tau)$ instead of gating actions, {one value for the agent and one for its environment}. The
difference from enforcement is that the trace is fixed, so there is no
intervention; and because each valuation $\llbracket C_i \rrbracket$ depends only on the trace, on the same trace $\tau$ the offline replay computes the same prefix valuations $\llbracket C_i \rrbracket(u)$ as the online monitor, and its end verdict relates to them as stated in Proposition~\ref{prop:agree}. The same incremental monitor is used for every trace, so the offline cost is linear in the total number of events.
Offline evaluation is thus deterministic and reproducible: the same trace always yields the same verdict and per-contract attribution.

\section{Evaluation}
\label{sec:experiments}

\yifeng{Our implementation\footnote{\url{https://github.com/SponsioLabs/ContrAgent}} is in Python 3.12 and was originally derived from the open-source Sponsio runtime~\cite{sponsio2026}. It supports exporting contract libraries to CHASE~\cite{nuzzo2018chase} for design-time contract analysis. The implementation uses \texttt{mus2muc}~\cite{ielo2026ltlfmuc} for conflict-core extraction and Z3~\cite{de2008z3} for numeric consistency checking.}
{The contract library is built with an \gls{llm} following the formulation pipeline of \S\ref{sec:framework-compile}, over the interaction predicates of Table~\ref{tab:atoms}. Integration hooks intercept each event and ground it deterministically through pattern matches and counter updates.}
Latency measurements reported below were taken on an Apple M4 Pro laptop~(24\,GB).

\paragraph{Benchmarks.}
\yifeng{We evaluate the two supervision roles on four benchmarks. SOPBench~\cite{sopbench2025} tests live enforcement of operating procedures (\S\ref{sec:eval-sopbench}). AgentDojo~\cite{agentdojo2024} tests whether the monitor would block unsafe calls in recorded prompt-injection runs (\S\ref{sec:eval-agentdojo}). R-Judge~\cite{rjudge2024} tests offline safety-risk detection (\S\ref{sec:eval-rjudge}), and $\tau^2$-bench~\cite{barres2025tau2} tests offline procedural-compliance scoring alongside outcome \mbox{success (\S\ref{sec:eval-tau2})}.}

\subsection{Standard Operating Procedure Enforcement}
\label{sec:eval-sopbench}

SOPBench \cite{sopbench2025} evaluates whether a language agent follows an
explicit standard operating procedure (SOP) across seven customer-service domains
(\textsf{bank}, \textsf{DMV}, \textsf{healthcare}, \textsf{hotel}, \textsf{library}, \textsf{university}, \textsf{online market}), where each task
provides a constraint graph, numeric thresholds, an initial database, and a label indicating whether the policy permits the goal.
We compared \sysname with three baselines: (1) the base model with no SOP, (2) the base model prompted with the SOP, and (3) a second \gls{llm} that judges each call against the SOP.
We report \emph{success} and \emph{safety} metrics, representing goal completion on tasks the SOP permits and correct blocking on tasks the SOP forbids, respectively.

We build each domain's agent contracts from its public SOP, compiling the SOP's gate/chain tree.
As shown in Table~\ref{tab:sopbench-main}, \sysname enforcement keeps mean success close to the base level ($90\%$ vs.\ $91\%$) while raising safety from $32\%$ to $98\%$, above prompt's $94\%$. Prompting and \gls{llm}-guard also achieve high safety ($94\%$ and $98\%$), but reduce mean success to $64\%$ and $24\%$, respectively, because of over-blocking.
{Both residual gaps trace to the agent rather than the enforcement layer: in \textsf{healthcare}, where success drops from $90\%$ to $83\%$, the agent receives the block message and is allowed to retry but does not, and in \textsf{university} it makes a permitted change to a target the benchmark's outcome-based scoring cannot distinguish from the forbidden one.}

The runtime cost of \sysname is lower than both prompt and \gls{llm}-guard, since it neither increases the prompt length nor calls a second model on every action.
Each entry is a mean over {three} seeded trials; \sysname's enforcement is stable across trials, with per-domain deviation $\le 0.4$~\gls{pp}, versus $2$ to $4$~\gls{pp} for the model-driven conditions.

\begin{table}[t]
\centering
\small
\setlength{\tabcolsep}{5pt}
\renewcommand{\arraystretch}{1.1}
\begin{tabular}{@{}lcccc@{}}
\toprule
\textbf{domain} {\scriptsize{(Su./Sf.)}} & \textbf{base} & \textbf{prompt} & \textbf{\gls{llm}-grd} & \textbf{\sysname} \\
\hline
\textsf{bank}          & 70/65  & 58/100 & 42/100 & \textbf{70/100} \\
\textsf{DMV}           & 100/60 & 72/98  & 20/100 & \textbf{100/100} \\
\textsf{healthcare}    & 90/48  & 52/95  & 20/100 & \textbf{83/100} \\
\textsf{hotel}         & 100/0  & 30/98  & 15/100 & \textbf{100/100} \\
\textsf{library}       & 80/25  & 48/95  & 22/98 & \textbf{80/100} \\
\textsf{university}    & 100/8  & 92/82  & 33/88 & \textbf{100/83} \\
\textsf{online mkt.}   & 100/20 & 92/90  & 15/98 & \textbf{100/100} \\
\hline
\textbf{mean} & 91/32 & 64/94 & 24/98 & \textbf{90/98} \\
\hline
\textbf{avg. runtime} & $1.96$\,s & $+0.905$\,s & $+1.34$\,s & $+0.135$\,s \\
\bottomrule
\end{tabular}
\caption{SOPBench results (\texttt{gemini-2.5-flash}, 40 tasks/domain); {each cell is success (Su.)\,/\,safety (Sf.) \%}.
\textbf{avg. runtime}: the \textbf{base} cell is its median runtime per
task; the other cells are the median \emph{extra} time over base. 
}
\label{tab:sopbench-main}
\end{table}

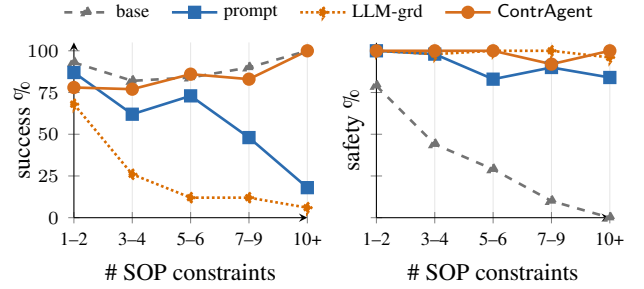
\begin{figure}[t]
\centering
\begin{tikzpicture}
\begin{groupplot}[
  group style={group size=2 by 1, horizontal sep=26pt},
  width=0.555\columnwidth, height=3.9cm,
  xlabel={\# SOP constraints},
  xtick={1,2,3,4,5}, xticklabels={1--2,3--4,5--6,7--9,10+},
  ymin=0, ymax=105, ytick={0,25,50,75,100},
  ylabel near ticks, ylabel shift=-8pt, ylabel style={font=\footnotesize},
  label style={font=\footnotesize}, tick label style={font=\scriptsize},
  xmajorgrids, grid style={line width=.2pt, draw=gray!20}, axis lines=left,
]
\nextgroupplot[ylabel={success \%},
  legend style={font=\scriptsize, legend columns=4, at={(1.10,1.06)}, anchor=south,
    draw=none, fill=none, /tikz/every even column/.append style={column sep=6pt}}]
\addplot[black!55, mark=triangle*, line width=1pt, dashed] coordinates {(1,93)(2,82)(3,84)(4,90)(5,100)};
\addplot[cgblue, mark=square*, line width=1pt] coordinates {(1,87)(2,62)(3,73)(4,48)(5,18)};
\addplot[orange!85!black, mark=diamond*, line width=1pt, densely dotted] coordinates {(1,68)(2,26)(3,12)(4,12)(5,6)};
\addplot[cgaccent, mark=*, line width=1pt] coordinates {(1,78)(2,77)(3,86)(4,83)(5,100)};
\legend{base, prompt, \gls{llm}-grd, \sysname}
\nextgroupplot[ylabel={safety \%}, yticklabels={,,}]
\addplot[black!55, mark=triangle*, line width=1pt, dashed] coordinates {(1,79)(2,44)(3,29)(4,10)(5,0)};
\addplot[cgblue, mark=square*, line width=1pt] coordinates {(1,100)(2,98)(3,83)(4,90)(5,84)};
\addplot[orange!85!black, mark=diamond*, line width=1pt, densely dotted] coordinates {(1,100)(2,98)(3,100)(4,100)(5,96)};
\addplot[cgaccent, mark=*, line width=1pt] coordinates {(1,100)(2,100)(3,100)(4,92)(5,100)};
\end{groupplot}
\end{tikzpicture}
\caption{SOPBench success (left) and safety (right) vs.\ SOP constraint count,
pooled over seven domains.}
\label{fig:sopbench-scaling}
\end{figure}

\begin{figure*}[t]
\centering
\begin{minipage}[b]{0.64\textwidth}
\centering
{\footnotesize
\raisebox{0.25ex}{\tikz{\fill[black!70] (0,0) circle (1.5pt);}}\,base \quad
\raisebox{0.25ex}{\tikz{\draw[cgaccent, line width=0.8pt, fill=white] (0,0) circle (1.5pt);}}\,{no detector} \quad
\raisebox{0.25ex}{\tikz{\draw[cgblue, line width=0.8pt, fill=white] (0,0) circle (1.5pt);}}\,trace-learned}\\[2pt]
\begin{tikzpicture}
\begin{groupplot}[
  group style={group size=2 by 1, horizontal sep=80pt},
  width=0.36\textwidth, height=4.7cm,
  xmode=log, xmin=0.08, xmax=70, xtick={0.1,1,10}, log ticks with fixed point,
  y tick label style={font=\ttfamily\scriptsize},
  tick label style={font=\footnotesize},
  xmajorgrids, grid style={line width=.2pt, draw=gray!20},
  clip=false, enlarge y limits=false,
]
\nextgroupplot[axis x line=bottom, axis y line=left, ymin=11.4, ymax=22.6,
  ytick={12,13,14,15,16,17,18,19,20,21,22},
  yticklabels={gemini-2-flash, gemini-2-flash-exp, gemini-1.5-pro-2, llama-3-70b,
    claude-3-sonnet, gpt-4o-mini, gemini-1.5-pro, gpt-4-turbo, gpt-4o,
    claude-3.5-sonnet, gpt-4-0125}]
\addplot[draw=none, mark=none] coordinates {(0.1,12)};
\foreach \yy/\base/\guard/\learn in {%
  22/56.3/15.6/4.5, 21/33.9/12.9/3.5, 20/28.6/8.2/3.0, 19/28.6/6.2/0.3, 18/28.6/10.0/5.6,
  17/27.2/6.4/0.8, 16/26.7/8.3/1.4, 15/25.6/13.2/6.2, 14/17.0/4.5/1.0, 13/17.0/4.6/1.4,
  12/14.1/2.7/0.4}{%
  \edef\db{%
    \noexpand\draw[gray!70, line width=1.0pt] (axis cs:\learn,\yy) -- (axis cs:\base,\yy);%
    \noexpand\fill[black!70] (axis cs:\base,\yy) circle (1.6pt);%
    \noexpand\draw[cgaccent, line width=0.8pt, fill=white] (axis cs:\guard,\yy) circle (1.6pt);%
    \noexpand\draw[cgblue, line width=0.8pt, fill=white] (axis cs:\learn,\yy) circle (1.6pt);%
  }%
  \db
}
\nextgroupplot[axis x line=bottom, axis y line=left, ymin=0.4, ymax=11.6,
  ytick={1,2,3,4,5,6,7,8,9,10,11},
  yticklabels={claude-3.5-sonnet-2, meta-secalign-70b, command-r, gemini-1.5-flash-2,
    command-r-plus, claude-3.7-sonnet, claude-3-haiku, gpt-3.5-turbo, claude-3-opus,
    gemini-1.5-flash, llama-3.3-70b}]
\addplot[draw=none, mark=none] coordinates {(0.1,1)};
\foreach \yy/\base/\guard/\learn in {%
  11/12.9/2.4/0.5, 10/12.2/4.1/1.7, 9/11.3/3.8/1.4, 8/10.3/3.5/2.1, 7/9.1/3.0/0.1,
  6/5.0/2.1/0.1, 5/4.5/3.7/3.5, 4/3.5/1.3/0.5, 3/3.3/1.9/0.5, 2/1.7/0.2/0.1,
  1/1.1/0.8/0.1}{%
  \edef\db{%
    \noexpand\draw[gray!70, line width=1.0pt] (axis cs:\learn,\yy) -- (axis cs:\base,\yy);%
    \noexpand\fill[black!70] (axis cs:\base,\yy) circle (1.6pt);%
    \noexpand\draw[cgaccent, line width=0.8pt, fill=white] (axis cs:\guard,\yy) circle (1.6pt);%
    \noexpand\draw[cgblue, line width=0.8pt, fill=white] (axis cs:\learn,\yy) circle (1.6pt);%
  }%
  \db
}
\end{groupplot}
\end{tikzpicture}\\[1pt]
{\footnotesize Prompt-injection attack-success rate (\%, log scale)}
\caption{Indirect prompt-injection prevention across $22$ \glspl{llm} on AgentDojo;
left and right are the higher- and lower-ASR halves.}
\label{fig:agentdojo22}
\end{minipage}
\hfill
\begin{minipage}[b]{0.34\textwidth}
\centering
{\footnotesize
\raisebox{0.25ex}{\tikz{\fill[black!70] (0,0) circle (1.5pt);}}\,pass\smash{$^4$} \quad
\raisebox{0.25ex}{\tikz{\draw[cgaccent, line width=0.8pt, fill=white] (0,0) circle (1.5pt);}}\,joint\smash{$^4$}}\\[2pt]
\begin{tikzpicture}
\begin{axis}[
  width=0.76\textwidth, height=4.7cm,
  xmin=0, xmax=64, ymin=-1.9, ymax=13.7,
  xtick={0,20,40,60},
  ytick={-1,0,1,2,4,5,6,7,9,10,11,12},
  yticklabels={o4-mini,GPT-4.1-mini,GPT-4.1,Claude 3.7,%
               o4-mini,GPT-4.1-mini,GPT-4.1,Claude 3.7,%
               o4-mini,GPT-4.1-mini,GPT-4.1,Claude 3.7},
  tick label style={font=\footnotesize},
  y tick label style={font=\ttfamily\scriptsize},
  label style={font=\small}, axis lines=left,
  xmajorgrids, grid style={line width=.2pt, draw=gray!20},
  clip=false, enlarge y limits=false,
]
\addplot[draw=none, mark=none] coordinates {(0,-1)};
\foreach \yy/\pa/\jo in {%
  12/59.6/0.0, 11/52.6/46.5, 10/38.6/13.2, 9/45.6/41.2,
  7/36.0/0.0,  6/40.0/28.0,  5/26.0/14.0, 4/38.0/30.0,
  2/25.4/0.0,  1/19.3/12.3,  0/17.5/2.6,  -1/26.3/12.3}{%
  \edef\dumbbell{%
    \noexpand\draw[gray!70, line width=1.1pt]
      (axis cs:\jo,\yy) -- (axis cs:\pa,\yy);%
    \noexpand\fill[black!70] (axis cs:\pa,\yy) circle (1.8pt);%
    \noexpand\draw[cgaccent, line width=0.8pt, fill=white]
      (axis cs:\jo,\yy) circle (1.8pt);%
  }%
  \dumbbell
}
\node[anchor=west, font=\footnotesize\bfseries] at (axis cs:0,13.2) {Retail};
\node[anchor=west, font=\footnotesize\bfseries] at (axis cs:0,8.1)  {Airline};
\node[anchor=west, font=\footnotesize\bfseries] at (axis cs:0,3.1)  {Telecom};
\end{axis}
\end{tikzpicture}\\[1pt]
{\footnotesize Task-level success rate (\%, $k=4$)}
\caption{The $\tau^2$-bench procedural reliability
across three domains and four models.}
\label{fig:tau2tax}
\end{minipage}
\end{figure*}

\paragraph{Model-agnostic supervision.}
\label{sec:eval-robust}
With the same contract library, on \texttt{gemini-2.5-flash-lite}, where the unguarded base stays at $27\%$, prompt's mean safety {collapses} from $94\%$ to $45\%$, whereas \sysname's enforced safety is {unchanged} at $98\%$,
because the verdict does not depend on the agent's model.
Moreover, replaying the same contract library over SOPBench traces from
$23$ base models, the false-positive rate on safe traces stays below $1\%$ for $19$ of $23$ models (max $4.5\%$, Llama-3.1-8B), even as the per-model violation rate varies widely with capability (Appendix~\ref{app:permodel}).

\paragraph{SOP constraint scaling.}
Binning tasks by the number of SOP constraints (which run from $1$ to $17$;
pooled over the seven domains) exposes the success/safety tradeoff that a single strategy hides. 
As shown in Figure~\ref{fig:sopbench-scaling}, with the number of constraints increasing, prompting increasingly over-refuses: its success on permitted tasks collapses from $87\%$ to $18\%$ in the $10+$ bin, while the unguarded base's safety collapses from $79\%$ to $0\%$. The \gls{llm}-guard keeps safety high, but with steep success reduction. \sysname holds success at $77\%$ or above and safety at \yifeng{$92\%$} or above across all bins, and the gap widens as the constraint count grows. Exact per-bin values and the weak-model results are in Appendix~\ref{app:sopbench}.

\subsection{Indirect Prompt Injection Prevention}
\label{sec:eval-agentdojo}

AgentDojo \cite{agentdojo2024} embeds attacker text inside tool outputs (email
bodies, calendar entries, search results) and measures whether the agent is
steered into an unsafe action.
\yifeng{We assess enforcement by replaying recorded runs and checking each proposed call before its recorded execution. The monitor uses tool-output provenance to block subsequent calls that violate the contract, regardless of whether the agent was tricked into issuing them. The replay evaluates the original trajectory up to the first block; it does not simulate the agent's response to that intervention.}
The contract library is built from the task and environment specifications of AgentDojo's four suites, without knowledge of the attacks.
It tags any value carried in by a tool output as untrusted, and blocks a side-effecting call whose target is both untrusted-introduced and outside the task's legitimate set. The legitimate set is read from each task's reference side effects and the suite's default environment. The untrusted-introduced condition is grounded by value matching. A target is untrusted if its normalized value, or a domain it contains, appears in a tool output and not in the user turn. This covers copied values and decorated or concatenated values that keep the value or domain intact.

Figure~\ref{fig:agentdojo22} reports three results per model. The baseline results are from AgentDojo's published numbers~\cite{agentdojo2024}.
{We call \emph{no-detector} the setting in which no component flags the injected text, so that a value counts as untrusted whenever a tool output introduces it and the user has not named it. In this setting, the library reduces the pooled \gls{asr} from $18.0\%$ to $5.3\%$ at $0.8\%$ utility \gls{fp}.}
Contracts can also be mined from the recorded traces of the attacks themselves {(attacker recipients, unsafe actions, and the injected text as the untrusted source)}, which we call \emph{trace-learned} contracts.
\yifeng{This lowers the pooled \gls{asr} to $1.7\%$ at $0.3\%$ \gls{fp}. Trace-learned contracts are mined from the same attacks they are scored on, with no held-out split, so this setting is an in-sample diagnostic. The no-detector library is the attack-agnostic result; it still uses the task reference information described above to define legitimate targets.}
The attacks \sysname misses fall into two kinds.
Most ($82\%$) issue no malicious tool call at all, e.g., denial-of-service injections that only divert the agent or text-response attacks answered in the agent's own reply, outside what a tool-call shield can do.
The remaining $18\%$ are prompt injections that steer the agent into a tool call whose arguments are themselves legitimate values, leaving the deterministic guard no untrusted target to flag.

\begin{table}[t]
\centering
\footnotesize
\setlength{\tabcolsep}{4pt}
\renewcommand{\arraystretch}{1.25}
\begin{tabular}{@{}llrrr@{}}
\toprule
\textbf{Defense} & \textbf{Type} & \textbf{ASR} & \textbf{Utility} & \textbf{{Overhead}} \\
\hline
No defense              & raw agent        & 47.7\% & 72.6\% & 0 \\
spotlighting            & prompt-side      & 41.7\% & 75.0\% & prompt \\
repeat\_user\_prompt    & prompt-side      & 27.8\% & 85.5\% & prompt \\
\textbf{\sysname (ND)} & provenance & \textbf{11.1\%} & \textbf{71.8\%} & \textbf{0.16 ms} \\
pi\_detector            & classifier       & 7.95\% & 41.9\% & ${\sim}50$ ms \\
tool\_filter            & \gls{llm} prune  & 6.84\% & 71.8\% & ${\sim}500$ ms \\
LlamaFirewall           & guardrail        & 1.75\% & n/a    & ${\sim}100$ ms \\
\textbf{\sysname (TL)} & provenance & \textbf{0.79\%} & \textbf{72.6\%} & \textbf{0.16 ms} \\
\bottomrule
\end{tabular}
\caption{AgentDojo defenses on \texttt{gpt-4o} (ND: no detector, TL: trace-learned).}
\label{tab:agentdojo}
\vspace{-2mm}
\end{table}

Table~\ref{tab:agentdojo} compares \sysname with the defenses bundled with
AgentDojo on \verb|gpt-4o| under its main injection attack. Only the \sysname rows are measured here; the baseline rates are AgentDojo's published numbers{, and every utility value is computed from the published runs. Utility is the fraction of injection-free tasks completed. Overhead is the per-call runtime added by a defense, and prompt-side defenses add prompt text rather than a call. \sysname replays the recorded runs, so its utility cannot exceed that of the undefended agent. Replay is exact on every trace without a block and on every prefix up to the first block. After a block, the \gls{asr} counts the attack as stopped and utility counts the task as failed. With no detector, \sysname lowers the \gls{asr} from $47.7\%$ to $11.1\%$ and completes $71.8\%$ of the injection-free tasks, one task fewer than the undefended agent. The classifier reaches $7.95\%$ but completes $41.9\%$ of the tasks. The trace-learned variant reaches $0.79\%$ at the utility of the undefended agent. Both variants add $0.16$ ms per call, against $50$ to $500$ ms for the classifier and the \gls{llm} filter} (Appendix~\ref{app:latency}).

\subsection{Agent Safety Risk Evaluation}
\label{sec:eval-rjudge}

R-Judge~\cite{rjudge2024} tests whether an evaluator can detect unsafe agent behavior \emph{post hoc} across ten operational risk types in $571$ multi-turn agent records ($301$ unsafe / $270$ safe).
Contracts come from R-Judge's risk taxonomy, a policy document (\S\ref{sec:framework-compile}).
\sysname reaches an average F$_1$ score of $91.8\%$ ($97.0\%$ precision, $87.0\%$ recall) across multiple trials.
This exceeds GPT-4o ($74.4\%$) and the R-Judge human baseline ($\approx\!89\%$).
AgentAuditor~\cite{agentauditor2025} outperforms our method at the cost of a Gemini-2 judge with retrieval over an experiential memory.
The records we miss are semantic rather than procedural, outside what a deterministic tool-call check can decide.

\begin{figure}[t]
\centering
% Figure: per-contract-group violation shares on tau2-bench (TikZ, no external file)
\begin{tikzpicture}[x=1cm,y=1cm,font=\scriptsize]
\def\cw{0.50}\def\ch{0.56}\def\gap{0.20}\def\lx{2.05}
% domain headers and model labels
\foreach \d/\dname in {0/Retail,1/Airline,2/Telecom}{
  \pgfmathsetmacro\gx{\lx+\d*(4*\cw+\gap)}
  \node[font=\scriptsize\bfseries,anchor=south] at (\gx+2*\cw,0.02) {\dname};
  \foreach \c/\m in {0/Cl,1/G,2/Gm,3/o4}{
    \node[anchor=north] at (\gx+\c*\cw+0.5*\cw,-4*\ch-0.02) {\m};}
}
% cells
\foreach \r/\vals in {0/{99,2,33,0,97,12,39,0,99,21,58,0},
                      1/{0,0,0,0,12,8,9,6,0,0,14,0},
                      2/{1,3,5,3,6,5,4,3,20,14,17,46},
                      3/{0,1,1,0,0,0,0,2,0,0,0,1}}{
  \foreach [count=\c from 0] \v in \vals {
    \pgfmathsetmacro\gi{int(\c/4)}
    \pgfmathsetmacro\cx{\lx+\c*\cw+\gi*\gap}
    \pgfmathsetmacro\cy{-\r*\ch}
    \pgfmathsetmacro\pct{min(100,\v)}
    \pgfmathsetmacro\tc{\v>=55 ? 1 : (\v>0 ? 2 : 3)}
    \fill[cgblue!\pct!white] (\cx+0.03,\cy-0.03) rectangle (\cx+\cw-0.03,\cy-\ch+0.03);
    \ifnum\tc=1 \node[text=white] at (\cx+0.5*\cw,\cy-0.5*\ch) {\v};
    \else\ifnum\tc=2 \node at (\cx+0.5*\cw,\cy-0.5*\ch) {\v};
    \else \node[text=gray!80] at (\cx+0.5*\cw,\cy-0.5*\ch) {\v};\fi\fi
  }
}
% row labels
\foreach \r/\lab in {0/{Text and tool call\\[-2.5pt]in one turn},1/{Argument value},2/{Look-up before\\[-2.5pt]action},3/{Call limit}}{
  \node[anchor=east,align=right,inner sep=0pt] at (\lx-0.12,-\r*\ch-0.5*\ch) {\lab};}
\end{tikzpicture}
\vspace{-2mm}
\caption{Share (\%) of $\tau^2$-passing traces that violate each contract group (Cl: Claude~3.7, G: GPT-4.1, Gm: GPT-4.1-mini, o4: o4-mini).}
\label{fig:tau2contract}
\end{figure}

\subsection{Procedural Violation and Outcome Scoring}
\label{sec:eval-tau2}

$\tau^2$-bench \cite{barres2025tau2} runs customer-service agents through
multi-turn dual-control conversations against a written policy document in
three domains. Its native \texttt{pass}$^k$ metric is outcome-only (final
database state). 
AgentPex \cite{sharma2026willful} showed $83\%$ of reward-$1.0$ Claude traces still violate a procedural rule. 
With contract libraries built from the policy documents, \sysname measures procedural compliance deterministically, while \gls{llm} judges can only approximate it.
We report a \texttt{joint}$^k$ metric that combines outcome and zero contract violations across $k$ retries.
Figure~\ref{fig:tau2tax} exposes the gap this creates: across the $4{,}464$-trace matrix Claude~3.7 reaches an outcome \texttt{pass}$^4$ of $25$--$60\%$ yet a \texttt{joint}$^4$ of $0\%$ in all three domains, i.e., no task is completed reliably without at least one procedural violation, a pattern \citet{cao2026corrupt} call corrupt success. The library holds the $37$ policy rules that are decidable from tool names, arguments, and call order, and omits rules that need facts from the environment database. Figure~\ref{fig:tau2contract} reports, for each rule group, the share of $\tau^2$-passing traces that violate it. For Claude~3.7, $97$--$99\%$ of the passing traces contain both user-facing text and a tool call in the same turn, which all three policies forbid. When this rule is excluded, its \texttt{joint}$^4$ rises to $59\%$, $26\%$, and $19\%$. The other models show different violations. For example, in the telecom domain, o4-mini often takes an action before looking up the data usage ($46\%$). Per-cell values for every domain and model are in Appendix~\ref{app:tau2}.

\paragraph{Limitations.}
\sysname enforces only what the tool-call structure exposes, and
free-form semantic harms need a separate content-level check.
\yifeng{The experiments use libraries authored with \gls{llm} assistance and inspected by humans before use. We evaluate supervision with these reviewed libraries; establishing the correctness of fully automatic policy formalization remains future work.} {The counters that ground numeric predicates (\S\ref{sec:contracts})
trust the framework to report events faithfully, and defending this
observation channel itself is beyond our scope.} \yifeng{The provenance check in \S\ref{sec:eval-agentdojo} matches values but does not track model-rewritten~information.}

\section{Conclusion}
\label{sec:conclusion}

We presented \sysname, a contract-based framework that formalizes
trajectory-level specifications of large language model (LLM) agents as
assume-guarantee (A/G) contracts over checkable tool-call predicates,
expressed in arithmetic linear temporal logic on finite traces
(ALTL$_f$), and compiles each into a deterministic finite automaton
(DFA) for online enforcement and offline evaluation.
\yifeng{The contract library provides reusable agent policies independently of the backbone model. Across four benchmarks, we evaluated live procedure enforcement, replay-based intervention checks, safety-risk detection, and procedural-compliance scoring. The results show that \sysname is competitive with LLM-judge, classifier, and guardrail baselines while producing deterministic, reproducible verdicts, with orders of magnitude lower per-call checking latency than model-based guards.}
{Future work includes compiling contracts into reward machines
for post-training and constrained decoding, and investigating contract-based decomposition mechanisms for multi-agent systems, where the split of agent and environment variables supports the composition and refinement of contracts \yifeng{across~agents}.}

\bibliography{reference}

\clearpage
\appendix
\onecolumn
\section{Full Interaction-Predicate Catalogue}
\label{app:atoms}

Table~\ref{tab:atoms-full} lists the complete interaction-predicate vocabulary of \sysname. The main text (Table~\ref{tab:atoms}) shows the representative subset that the examples and experiments use; the remaining predicates extend the same two families to model outputs, response lengths, and delegation depth. $\mathsf{Since}(a)$ reads a wall-clock timestamp the framework records with each event; time is an implementation-level extension and does not enter the formal model of Def.~\ref{def:session}.

\begin{table}[h]
\centering
\footnotesize
\setlength{\tabcolsep}{6pt}
\renewcommand{\arraystretch}{1.2}
\begin{tabularx}{\textwidth}{@{}c l X@{}}
\toprule
\textbf{Type} & \textbf{Predicate} & \textbf{Meaning} \\
\midrule
\multirow{11}{*}{\rotatebox[origin=c]{90}{{Structural}}}
  & $\mathsf{Call}(T)$ & tool $T$ is invoked \\
  & $\mathsf{ArgHas}(T,f,p)$ & argument $f$ of $T$ matches pattern $p$ \\
  & $\mathsf{Path}(T,Q)$ & $T$'s file paths lie within $Q$ \\
  & $\mathsf{Subset}(f,S)$ & values in field $f$ lie within set $S$ \\
  & $\mathsf{OutHas}(T,p)$ & result of $T$ matches pattern $p$ \\
  & $\mathsf{Said}(p)$, $\mathsf{In}(p)$ & model output / input matches pattern $p$ \\
  & $\mathsf{Match}(f,k)$ & field $f$ equals the value of context key $k$ \\
  & $\mathsf{Ctx}(k,w)$ & context key $k$ holds value $w$ \\
  & $\mathsf{Flow}(g,g')$ & agent $g'$ reads data written by agent $g$ \\
  & $\mathsf{Has}(f)$ & a produced value contains field $f$ \\
  & $\mathsf{Perm}(Q)$ & caller holds permission $Q$ \\
\midrule
\multirow{9}{*}{\rotatebox[origin=c]{90}{{Numeric}}}
  & $\mathsf{Cnt}(T)$ & current number of $T$ calls \\
  & $\mathsf{Run}(T)$ & length of the current consecutive run of $T$ \\
  & $\mathsf{Num}(T,f)$ & numeric value of argument field $f$ \\
  & $\mathsf{Len}(T,f)$ & character length of argument field $f$ \\
  & $\mathsf{InLen}$ & character length of the model input \\
  & $\mathsf{Words}$, $\mathsf{Chars}$ & word / character length of the response \\
  & $\mathsf{Tok}$ & cumulative tokens consumed \\
  & $\mathsf{Depth}$ & agent-delegation depth \\
  & $\mathsf{Since}(a)$ & time elapsed since event $a$ \\
\bottomrule
\end{tabularx}
\caption{Complete interaction-predicate vocabulary of \sysname. $T$ is a tool, $f$ an argument field, $p$ a regular-expression or literal-value pattern, $Q$ a path or permission set, $S$ a value set, $g,g'$ agents, and $k,w$ a context key/value.}
\label{tab:atoms-full}
\end{table}

\section{Temporal Expressiveness}
\label{app:temporal}

Table~\ref{tab:temporal} lists four temporal property classes that distinguish
\sysname's trajectory-level enforcement from a stateless, single-call guard
(\S\ref{sec:contracts}). For each we construct
a minimal violating trace in which \emph{every individual call is locally
legitimate} (the same calls occur in a compliant trace), so the violation is
purely temporal. \sysname's \gls{dfa} catches all four and raises no false
positive on the compliant control; a stateless guard catches none by construction. The
fourth class, ``after reading untrusted content, a side-effecting send requires
reconfirmation,'' is exactly the indirect-prompt-injection contract that drives
the AgentDojo and R-Judge results (\S\ref{sec:eval-agentdojo},
\S\ref{sec:eval-rjudge}).

\begin{table}[h]
\centering
\footnotesize
\setlength{\tabcolsep}{3pt}
\renewcommand{\arraystretch}{1.25}
\begin{tabular}{@{}p{0.23\columnwidth}p{0.37\columnwidth}cc@{}}
\hline
\textbf{Property (NL)} & \textbf{ALTL$_f$} & \textbf{state\-less} & \textbf{\sysname} \\
\hline
Refund only after a policy check (order)
 & $(\neg\,\mathsf{Call}(\mathsf{r})\,U\,\mathsf{Call}(\mathsf{c}))\,\lor\,G\,\neg\,\mathsf{Call}(\mathsf{r})$
 & miss & \textbf{catch} \\
After the AML check, the file is immutable (history)
 & $G(\mathsf{Call}(\mathsf{a}) \rightarrow G\,\neg\,\mathsf{Call}(\mathsf{m}))$
 & miss & \textbf{catch} \\
At most two transfers (count)
 & $G(\mathsf{Cnt}(\mathsf{t})\leq 2)$
 & miss & \textbf{catch} \\
Send needs reconfirm after untrusted read (injection)
 & $(\neg\,\mathsf{Call}(\mathsf{s})\,U\,\mathsf{Call}(\mathsf{k}))\,\lor\,G\,\neg\,\mathsf{Call}(\mathsf{s})$
 & miss & \textbf{catch} \\
\hline
\end{tabular}
\caption{Temporal expressiveness: each violating trace's every call is
individually legitimate, so the violation is detectable only from order, history,
or count. A stateless guard catches none; \sysname's \gls{dfa} catches all four,
clean on the compliant control. Tool names in the formulas are abbreviated per row.}
\label{tab:temporal}
\end{table}

\section{Per-Model Violation Profiles and Grounding Ablation}
\label{app:permodel}

This appendix expands the grounding and per-model robustness summary in
\S\ref{sec:eval-robust}.

\paragraph{The residual gap is coverage, not grounding.}
SOPBench's decisive predicates are \emph{structural} (a tool was called, a call
returned success), read directly off the trace, so grounding is exact by
construction. We confirm this with an oracle ablation over $5{,}435$ recorded
unsafe traces: perfecting the single most influential \gls{ap}, the goal
action's success flag, moves pooled detection recall from $75.1\%$ to $75.4\%$
(a $+0.3$~\gls{pp} delta). There is essentially no grounding error to remove; the
residual $\sim\!25\%$ is contract \emph{coverage} (\S\ref{sec:eval-robust}). Oracle grounding thus matters for content and semantic sensors (e.g., R-Judge's semantic residual cases,
\S\ref{sec:eval-rjudge}), not for structural enforcement.

\paragraph{Per-model violation profiles.}
Table~\ref{tab:permodel} profiles the contract library across base models on
their recorded traces. The violation \emph{rate} tracks capability: the strongest
models almost never complete a forbidden action (gpt-5: $14$ unsafe traces in the
sample; o1: $3$), whereas mid-tier models do so routinely (Claude-3.5: $417$), a
direct, model-specific ``willful-disobedience'' measure. \sysname's behaviour on
those violations stays stable regardless: recall sits in the $54.7$--$100\%$ band and
the dominant family of caught violations is value/threshold for every model ($74$--$100\%$ of its
catches), at $<\!1\%$ false positives, so the supervision behavior does not depend on the backbone model.

\begin{table}[h]
\centering
\footnotesize
\setlength{\tabcolsep}{4pt}
\begin{tabular}{@{}lrrr@{}}
\hline
\textbf{base model} & \textbf{recall} & \textbf{FPR} & \textbf{n\textsubscript{uns}} \\
\hline
gpt-5                       & 100.0\% &  0.0\% &  14 \\
o1                          & 100.0\% &  0.0\% &   3 \\
gpt-5-mini                  &  92.9\% &  0.0\% &  42 \\
gemini-2.0-flash            &  81.0\% &  0.5\% & 420 \\
llama3.1-8b-instruct        &  79.3\% &  4.5\% & 416 \\
gemini-2.0-flash-thinking   &  79.2\% &  0.2\% & 101 \\
llama3.1-70b-instruct       &  79.0\% &  1.4\% & 415 \\
qwen2.5-72b-instruct        &  78.4\% &  0.5\% & 398 \\
o4-mini                     &  78.0\% &  0.0\% &  41 \\
qwen2.5-7b-instruct         &  77.9\% &  2.5\% & 408 \\
gpt-4o-mini                 &  77.3\% &  2.9\% & 299 \\
gemini-1.5-pro              &  75.9\% &  0.7\% & 332 \\
qwen2.5-14b-instruct        &  75.8\% &  0.0\% & 372 \\
qwen2.5-32b-instruct        &  75.2\% &  0.2\% & 359 \\
claude-3.5-sonnet           &  74.3\% &  0.0\% & 417 \\
gpt-4o                      &  74.2\% &  0.0\% & 322 \\
claude-3.7-sonnet           &  71.7\% &  0.5\% & 240 \\
gpt-4.1-mini                &  66.4\% &  0.0\% & 220 \\
claude-3.7-sonnet-thinking  &  66.1\% &  0.0\% & 189 \\
gpt-4.1                     &  63.7\% &  0.0\% & 179 \\
o4-mini-high                &  61.2\% &  0.2\% &  80 \\
deepseek-r1                 &  56.2\% &  0.3\% &  73 \\
gemini-2.5-flash            &  54.7\% &  0.0\% &  95 \\
\hline
\end{tabular}
\caption{Per-model violation profiles on SOPBench recorded traces (deterministic,
$0$ \gls{llm} calls; all $23$ base models). \textbf{recall} =
unsafe caught, \textbf{FPR} = safe wrongly blocked,
\textbf{n\textsubscript{uns}} = sampled unsafe traces. FPR stays below $1\%$ for
$19$ of $23$ backbones (max $4.5\%$, llama3.1-8b).}
\label{tab:permodel}
\end{table}

\section{SOPBench Live Enforcement: Per-Constraint Scaling and Robustness}
\label{app:sopbench}

Table~\ref{tab:scaling-exact} gives the exact per-bin values plotted in
Figure~\ref{fig:sopbench-scaling}: SOPBench success (on permitted tasks) and
safety (on forbidden tasks) as a function of the number of SOP constraints,
pooled over the seven domains, for the base, prompt, \gls{llm}-guard, and \sysname-enforce
conditions on \texttt{gemini-2.5-flash}. \sysname is the only condition that
stays high on \emph{both} axes across the whole range: prompt's success collapses
from $87\%$ to $18\%$ as the policy grows, and the unguarded base's safety
collapses from $79\%$ to $0\%$, while \sysname-enforce holds success
$\geq\!77\%$ and safety \yifeng{$\geq\!92\%$} throughout, with its safety matching or
exceeding prompt at every bin.

\begin{table}[h]
\centering
\footnotesize
\setlength{\tabcolsep}{5pt}
\begin{tabular}{@{}llrrrrr@{}}
\hline
\textbf{metric} & \textbf{condition} & \textbf{1--2} & \textbf{3--4} & \textbf{5--6} & \textbf{7--9} & \textbf{10+} \\
\hline
\multirow{4}{*}{success}
 & base             & 93  & 82 & 84  & 90 & 100 \\
 & prompt           & 87  & 62 & 73  & 48 & 18  \\
 & \gls{llm}-grd    & 68  & 26 & 12  & 12 & 6   \\
 & \textbf{\sysname}& \textbf{78} & \textbf{77} & \textbf{86} & \textbf{83} & \textbf{100} \\
\hline
\multirow{4}{*}{safety}
 & base             & 79  & 44 & 29  & 10 & 0  \\
 & prompt           & 100 & 98 & 83  & 90 & 84 \\
 & \gls{llm}-grd    & 100 & 98 & 100 & 100 & 96 \\
 & \textbf{\sysname}& \textbf{100} & \yifeng{\textbf{100}} & \textbf{100} & \yifeng{\textbf{92}} & \yifeng{\textbf{100}} \\
\hline
\end{tabular}
\caption{Exact per-bin values for Figure~\ref{fig:sopbench-scaling}: SOPBench
success (permitted) and safety (forbidden) \% vs.\ SOP constraint count, pooled
over the seven domains (\texttt{gemini-2.5-flash}). As the policy grows, prompt's
success and base's safety both collapse; \sysname-enforce stays high on both axes.}
\label{tab:scaling-exact}
\end{table}

\paragraph{Robustness to a weaker base model.}
The weak-model result in \S\ref{sec:eval-robust} reports the mean over domains;
the effect is consistent across domains. On the weaker \texttt{gemini-2.5-flash-lite} agent,
prompt's mean safety falls from $94\%$ to $45\%$ (the agent stops reliably
reading and obeying the prompted SOP), whereas \sysname-enforce reproduces its
strong-model safety ($98\%$, unchanged) because the monitor verdict is computed
deterministically from the trace, not inferred by the agent. Success drops for
every condition under the weaker agent (enforce mean $90\!\to\!68$), confirming
that success tracks model capability while \sysname's safety does not. This confirms the distinction between deterministic and probabilistic supervision. A prompted
SOP's assurance degrades with the model that carries it, whereas a compiled contract's
does not.

\section{Hot-Path Latency Breakdown}
\label{app:latency}

Table~\ref{tab:latency} gives the per-workload before-call latency of the online
verifier.

\begin{table}[h]
\centering
\footnotesize
\setlength{\tabcolsep}{3pt}
\begin{tabular}{@{}lrrrr@{}}
\hline
\textbf{Workload} & \textbf{C} & \textbf{p50} & \textbf{p95} & \textbf{p99} \\
\hline
Synthetic (1 contract)     & 1        & 0.0052 & --    & 0.012 \\
SOPBench (per call)        & 38 to 65 & 0.097  & 0.171 & 0.272 \\
AgentDojo (per call)       & 8 to 14  & 0.162  & 0.620 & 0.933 \\
R-Judge (per record)       & 7        & 0.040  & 0.085 & 0.141 \\
$\tau^2$-bench (per call)  & 43 & 0.592 & 1.067 & 1.380 \\
\hline
\end{tabular}
\caption{Hot-path before-call latency (ms) on the incremental verifier, including predicate grounding and contract evaluation. \textbf{C} is the number of contracts evaluated.}
\label{tab:latency}
\end{table}

\section{Per-Cell $\tau^2$-bench Reliability Matrix}
\label{app:tau2}

Figure~\ref{fig:tau2tax} plots the \texttt{pass}$^4$ and
\texttt{joint}$^4$ endpoints of the $\tau^2$-bench reliability matrix.
Table~\ref{tab:tau2matrix} gives the exact per-cell values, including
the intermediate \texttt{proc-clean}$^4$ column (all four retries had
zero \sysname rule fires). \texttt{pass}$^4$ is the native outcome
metric, \texttt{joint}$^4$ requires both outcome and procedure-clean on
every retry, and the \texttt{pass}$^4$-to-\texttt{joint}$^4$ gap quantifies the
loss of procedural reliability. All values are computed with the $37$ policy rules of \S\ref{sec:eval-tau2}.

\begin{table}[h]
\centering
\footnotesize
\setlength{\tabcolsep}{4pt}
\begin{tabular}{@{}llrrr@{}}
\hline
\textbf{Domain} & \textbf{Model} & \textbf{pass}$^4$ & \textbf{proc-clean}$^4$ & \textbf{joint}$^4$ \\
\hline
retail  & Claude 3.7   & 59.6 & \textbf{0.0}  & 0.0  \\
        & GPT-4.1      & 52.6 & 88.6 & 46.5 \\
        & GPT-4.1-mini & 38.6 & 23.7 & 13.2 \\
        & o4-mini      & 45.6 & 92.1 & 41.2 \\
\hline
airline & Claude 3.7   & 36.0 & \textbf{0.0}  & 0.0  \\
        & GPT-4.1      & 40.0 & 42.0 & 28.0 \\
        & GPT-4.1-mini & 26.0 & 14.0 & 14.0 \\
        & o4-mini      & 38.0 & 64.0 & 30.0 \\
\hline
telecom & Claude 3.7   & 25.4 & \textbf{0.0}  & 0.0  \\
        & GPT-4.1      & 19.3 & 16.7 & 12.3 \\
        & GPT-4.1-mini & 17.5 & 2.6  & 2.6  \\
        & o4-mini      & 26.3 & 14.9 & 12.3 \\
\hline
\end{tabular}
\caption{$\tau^2$-bench task-level reliability ($k=4$, percentages).
\texttt{pass}$^4$ is the native outcome metric (all four retries reach
the correct final state); \texttt{proc-clean}$^4$ requires zero
\sysname rule fires on all four retries; \texttt{joint}$^4$ requires
both. Claude~3.7 reaches \texttt{proc-clean}$^4 = 0\%$ in every domain,
so its \texttt{joint}$^4$ is 0\% despite a 25 to 60\% outcome
\texttt{pass}$^4$.}
\label{tab:tau2matrix}
\end{table}

\end{document}